\documentclass[sigconf, screen]{acmart}

\usepackage{setspace}
\usepackage{multirow}
\usepackage{amsmath}
\usepackage{booktabs}
\usepackage{enumitem}
\usepackage{colortbl}
\usepackage{subcaption}
\usepackage{xspace}
\usepackage{graphicx}
\usepackage[normalem]{ulem}
\usepackage{fontawesome}
\useunder{\uline}{\ul}{}
\usepackage{pifont}
\usepackage{algorithm, algorithmic}

\AtEndPreamble{
    \usepackage[capitalize]{cleveref}
    \crefname{section}{Sec.}{Secs.}
    \Crefname{section}{Section}{Sections}
    \Crefname{table}{Table}{Tables}
    \crefname{table}{Tab.}{Tabs.}
}
\AtBeginDocument{
  }
\setcopyright{acmlicensed}
\copyrightyear{2025}
\acmYear{2025}
\acmConference[Conference'25]{Make sure to enter the correct
  conference title from your rights confirmation email}{October 27--31,
  2025}{Dublin, Ireland}

\newcommand{\ie}{\emph{i.e.}}
\newcommand{\eg}{\emph{e.g.}}

\definecolor{bluecitecolor}{rgb}{0,0.36,0.69}
\begin{document}
\title{\includegraphics[height=5ex]{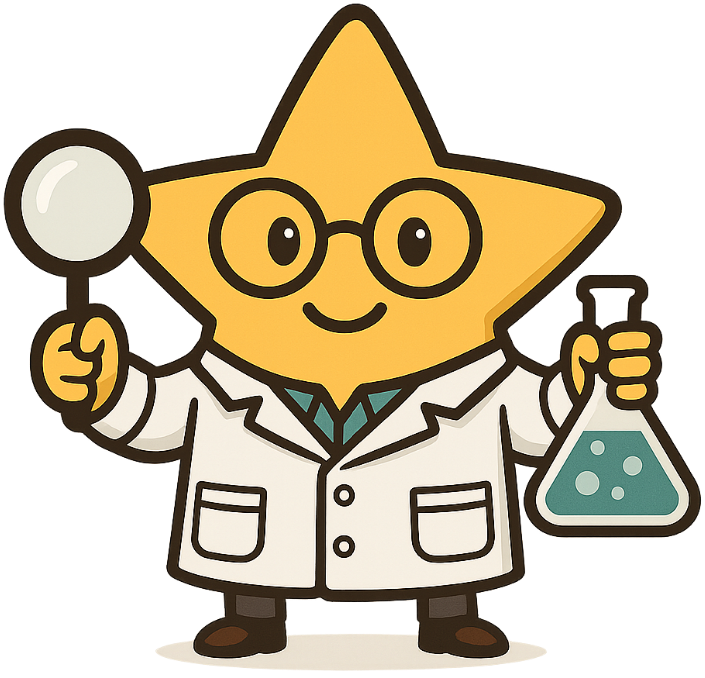}ExpStar: Towards Automatic Commentary Generation for Multi-discipline Scientific Experiments}

\author[J. Chen, Y. Jia, Z. Wu, J. Yang et al.]{
Jiali Chen$^{\heartsuit,\spadesuit*}$
\quad Yujie Jia$^{\heartsuit,*}$ 
\quad Zihan Wu$^{\heartsuit,\spadesuit}$ 
\quad Jinyu Yang$^{\heartsuit,\spadesuit}$ 
\quad Jianpeng Chen$^{\heartsuit}$
\quad Xusen Hei$^{\heartsuit,\spadesuit}$ \\
\quad Jiayuan Xie$^{\diamond}$
\quad Yi Cai$^{\spadesuit,\heartsuit,\dag}$
\quad Qing Li$^{\diamond}$
}

\affiliation{
\institution{\textsuperscript{\rm $\heartsuit$}South China University of Technology,
\textsuperscript{\rm $\diamond$}The Hong Kong Polytechnic University, \\
\textsuperscript{\rm $\spadesuit$}Key Laboratory of Big Data and Intelligent Robot Ministry of Education} 
\institution{\textbf{\href{https://cgl-pro.github.io/expstar}{\textcolor{bluecitecolor}{https://cgl-pro.github.io/expstar}}}}
}

\thanks{\textsuperscript{\rm *}Equal contribution.}
\thanks{\faEnvelope\ segarychen@mail.scut.edu.cn}
\thanks{\dag\ Correspondence}
\begin{abstract}

Experiment commentary is crucial in describing the experimental procedures, delving into underlying scientific principles, and incorporating content-related safety guidelines.
In practice, human teachers rely heavily on subject-specific expertise and invest significant time preparing such commentary.
To address this challenge, we introduce the task of automatic commentary generation across multi-discipline scientific experiments.
While recent progress in large multimodal models (LMMs) has demonstrated promising capabilities in video understanding and reasoning, their ability to generate fine-grained and insightful experiment commentary remains largely underexplored.
In this paper, we make the following contributions: (i) We construct \textit{ExpInstruct}, the first dataset tailored for experiment commentary generation, featuring over 7\textit{K} step-level commentaries across 21 scientific subjects from 3 core disciplines (\ie, science, healthcare and engineering). Each sample includes procedural descriptions along with potential scientific principles (\eg, chemical equations and physical laws) and safety guidelines.
(ii) We propose ExpStar, an automatic experiment commentary generation model that leverages a retrieval-augmented mechanism to adaptively access, evaluate, and utilize external knowledge. 
(iii) Extensive experiments show that our ExpStar substantially outperforms 14 leading LMMs, which highlights the superiority of our dataset and model.
We believe that ExpStar holds great potential for advancing AI-assisted scientific experiment instruction.

\end{abstract}

\renewcommand\footnotetextcopyrightpermission[1]{}

\settopmatter{printacmref=false} 
\renewcommand\footnotetextcopyrightpermission[1]{} 
\pagestyle{plain} 

\maketitle

\section{Introduction}
Scientific experiment instruction plays a critical role across various scientific subjects (\eg, physics and chemistry), extending beyond classroom teaching to serve as a bridge between theoretical knowledge and practical application \cite{sci-exp-1,sci-exp-2,sci-exp-3}.
During experiments, human teachers typically offer step-by-step commentary to guide students through the experimental procedures.
However, crafting pedagogically insightful and precise commentaries requires subject-specific expertise and a significant time investment \cite{sci-exp-3}. 
This constraint limits the scalability and consistency of high-quality experiment instruction across multiple disciplines.

\begin{figure}[!]
  \centering
  \includegraphics[scale=0.543]{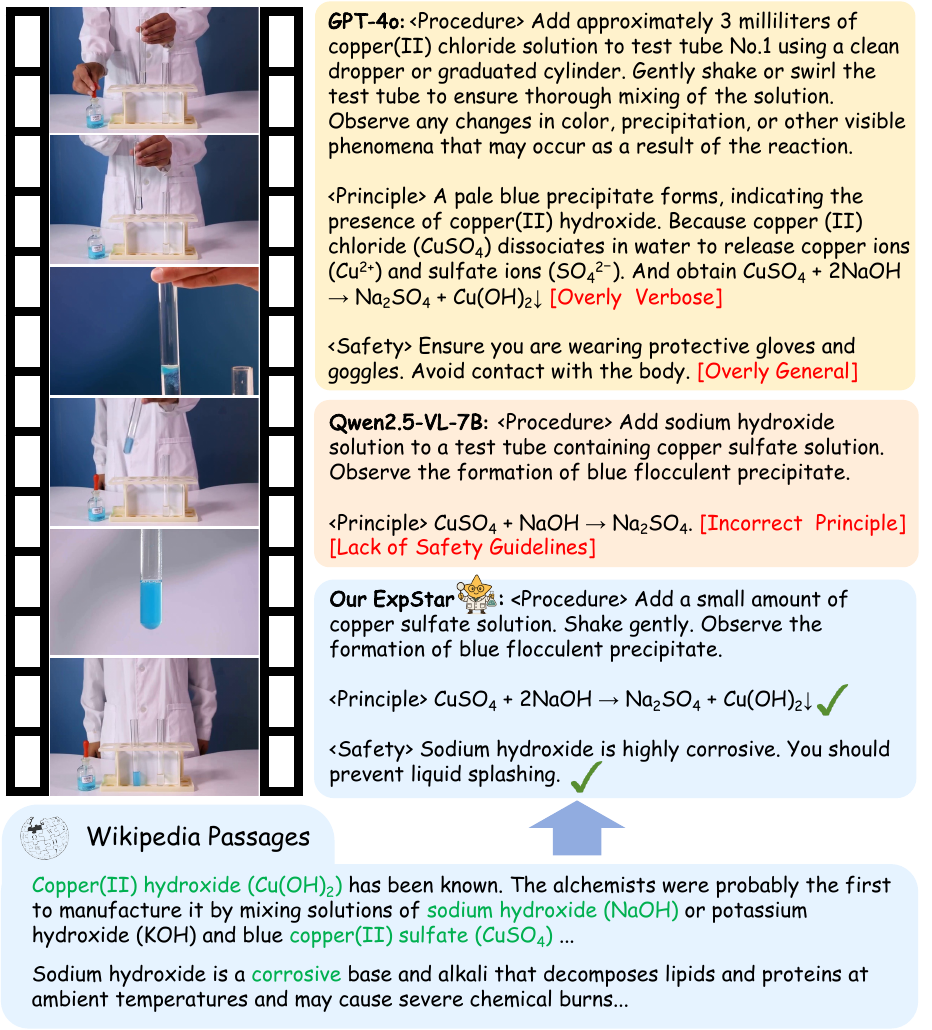}
  \caption{Generated commentaries by different models (\ie, GPT-4o, Qwen2.5-VL and our ExpStar) on a chemical reaction experiment between sodium hydroxide and copper sulfate.}
\label{fig: intro_case}
\end{figure}
Recent advances in large multimodal models (LMMs) have shown great promise in video understanding and reasoning \cite{mmworld,mmvu,videommmu}, which makes them a viable tool for generating experiment commentary.
Motivated by this insight, we introduce the experiment commentary generation task, which aims to automatically create commentaries for scientific experiment videos.
Scrutinizing existing LMMs in video captioning, they primarily focus on 
summarizing high-level events \cite{vid2seq,matchtime,streaming}.
Some recent studies \cite{cook,assem101} generate step-wise procedures about daily activities (\eg, cooking and equipment assembly), typically emphasizing operational sequences.
However, high-quality experiment commentaries should not only provide precise procedural descriptions but also delve into potential underlying scientific principles and incorporate content-related safety guidelines, as illustrated in Fig. \ref{fig: intro_case}.
Notably, GPT-4o tends to produce excessively verbose commentary that exceeds the temporal duration of the video, making it impractical for synchronized instructional use.
Furthermore, it generates overly generic safety instructions (\eg, ``avoid any contact'') that lack specificity for experimental procedures.

To tackle this challenge, we construct \textit{ExpInstruct}, the first multi-discipline dataset dedicated to scientific experiment commentary generation. 
It consists of over 7\textit{K} meticulously annotated step-level commentaries, covering 21 scientific subjects across 3 core disciplines (\ie, science, healthcare, and engineering). 
To ensure the high quality of commentaries and support experimental teaching, we design an AI-assisted annotation pipeline for dataset construction, complemented by rigorous human validation.
Specifically, we first collect diverse scientific experiment videos from publicly available online platforms and segment each video into distinct experimental steps (\ie, video clip).
For each step, we transcribe procedural descriptions with automatic speech recognition (ASR) tool (\ie, WhisperX \cite{whisperx}). Subsequently, we employ GPT-4o \cite{gpt4} to selectively annotate corresponding scientific principles and safety guidelines based on their contextual relevance.
Finally, expert annotators thoroughly verify all annotations, guaranteeing the superior quality of our \textit{ExpInstruct} dataset.
Leveraging constructed \textit{ExpInstruct} dataset, we develop \textbf{ExpStar}, an automatic experiment commentary generation model derived from the open-source LMM (\ie, Qwen2.5-VL-7B) with a retrieval-augmented mechanism.
It determines whether external knowledge retrieval is necessary and evaluates the relevance of retrieved information, thereby facilitating accurate and contextually appropriate experiment commentary generation. To achieve this, we augment ExpStar's vocabulary with additional special tokens designed to explicitly control the retrieval process.  
Specifically, ExpStar first generates experimental procedures based on the video clip. Then, it produces a retrieval control token indicating whether retrieval is necessary for principle or safety guideline generation. 
If retrieval is performed, ExpStar outputs relevance tokens to assess which of the retrieved Wikipedia passages are appropriate as relevant knowledge.
Following this paradigm, we employ supervised fine-tuning (SFT) to endow our ExpStar with the ability for adaptive knowledge retrieval and knowledge relevance judgment.
As illustrated in Fig. \ref{fig: intro_case}, relevant Wikipedia passages effectively support the generation of precise principles and safety guidelines.
Moreover, safety-related content is frequently overlooked by existing relatively small open-source LMMs, as evidenced by the absence of safety guidelines in the commentary generated by Qwen2.5-VL-7B in Fig. \ref{fig: intro_case}.
To address this shortcoming, we adopt a rule-based safety-aware reward optimization to refine our ExpStar. This refinement can improve the coverage and precision of safety guideline generation, ensuring that critical safety information is included accurately.

In summary, our contributions are as follows:
\begin{itemize}[leftmargin=*]
\item We construct \textit{ExpInstruct}, the first dataset for experiment commentary generation, featuring over 7\textit{K} step-level commentaries with corresponding video clips across 21 scientific subjects from 3 core disciplines (\ie, science, healthcare and engineering).
Each sample includes procedural descriptions along with potential scientific principles (\eg, chemical equations and physical laws) and safety guidelines.
\item We propose ExpStar, an automatic experiment commentary generation model that leverages a retrieval-augmented mechanism to adaptively access, evaluate, and utilize external knowledge.
This mechanism facilitates precise and contextually appropriate experiment commentary generation.
Moreover, ExpStar integrates a rule-based reward to explicitly encourage safety-related content generation. 
\item Extensive experiments demonstrate that our ExpStar achieves state-of-the-art performance, and significantly outperforms 14 leading LMMs. This highlights the superiority of our proposed dataset and model.
We believe our work will provide great potential for advancing AI-assisted experimental instruction.
\end{itemize}

\section{Related Work}
\subsection{Large Multimodal Models}
Large multimodal models (LMMs) have demonstrated remarkable multimodal reasoning \cite{mmmu,scienceqa,m3cot,l2c} and video understanding \cite{mvbench} abilities by integrating large language models (LLMs) \cite{llama-3,Qwen2} with vision encoders \cite{clip,siglip}.
Several LMMs (\eg, LLaVA\cite{llava,llavanext}, Qwen-VL \cite{qwen-vl,qwen2-vl,qwen2.5-vl}, Deepseek-VL ~\cite{deepseek-vl, deepseek-vl2} and InternVL \cite{internvl,intern-vl-2-5}) have emerged, which leverage larger, high-quality instruction datasets for fine-tuning. The frontier of LMMs has rapidly expanded into temporal dimensions, with video understanding emerging as a pivotal research direction \cite{apollo,videoxl}. 
Existing benchmarks (\eg, MM-World \cite{mmworld}, MMVU \cite{mmvu}, Video-MMMU \cite{videommmu}) have evaluated the video understanding capabilities of LMMs, focusing primarily on generic video captioning and video question answering. 
\subsection{Video Captioning}
Video captioning aims to automatically generate natural language descriptions for video content, transforming visual information into textual narratives \cite{dcaption, vid2seq}. 
Recent advances in video captioning have been driven by large multimodal models (LMMs) ~\cite{progress,sharegpt4video,captionlmm} that integrate visual understanding with language generation. LLaMA-VID \cite{llamavid} and Video-LLaMA \cite{videollama} introduce innovative strategies to connect visual features with language prompts. 
Notably, the field has witnessed the emergence of domain-specific video captioning models. The AutoAD series ~\cite{autoad,autoad2,autoad3} apply video captioning for movie scenes to assist visually impaired individuals in watching movies. \citet{cook} develop an automated captioning system for daily instructional videos (\eg, cooking). Automated soccer game commentary generation \cite{matchtime,soccer} enriches the viewing experience for audiences. 
Building on these advances, we present \textit{ExpInstruct}, the first high-quality multi-discipline scientific experiment commentary dataset. 
\section{Building \textit{ExpInstruct} Dataset}
In this section, we introduce the construction process of \textit{ExpInstruct} dataset.
Following previous benchmark datasets on video captioning \cite{soccer,goal}, \textit{ExpInstruct} consists of video-commentary pairs $(V, Y)$, where each video $V$ comprises sequential clips $\{v_1, v_2, \dots, v_n\}$ along with their corresponding commentaries $\{y_1, y_2, \dots, y_n\}$.
Notably, an informative experiment commentary should not only state the procedures, but also explain underlying scientific principles and provide safety reminders when necessary. In subsequent subsections, we first describe our source data collection (Sec. \ref{sec: data_col}), followed by data curation and manual quality control pipelines. (Sec. \ref{sec: data_cur} and Sec. \ref{sec: data_qua}). Finally, we present comprehensive statistics and analysis of our \textit{ExpInstruct} dataset (Sec. \ref{sec: data_ana}).

\begin{table}[]
\centering
\caption{\label{tab: sta} Key statistics of our \textit{ExpInstruct} dataset.}
\begin{spacing}{1.}
\resizebox{1.\columnwidth}{!}{
\begin{tabular}{l|c|ccc}
\toprule[1pt]
& \textbf{Total} & \textbf{Science} & \textbf{Healthcare} & \textbf{Engineering} \\ \midrule
Video Clips             & 7714   & 3565   & 2285       & 1864             \\
Average Duration       & 33.80   & 20.40   & 46.60     & 44.02         \\ 
Steps per Video        & 7.63 & 5.89  & 10.07  & 10.41         \\\midrule
Average Text Length     & 53.99   & 34.05         & 90.77     & 46.05         \\
\quad - Procedure                                                                                    & 36.78   & 23.40        & 67.58    & 24.60         \\
\quad - Principle                                                                                    & 19.34   & 14.66        & 23.00     & 21.08        \\
\quad - Safety Guideline                                                                                        & 16.89   & 17.90         & 17.42     & 14.29         \\ \midrule
Principle Rate                                                                                            & 52.04\%   & 41.65\%         & 60.79\%     & 62.35\%         \\
Safety Guideline Rate                                                                                           & 36.42\%   & 25.83\%         & 52.91\%     & 36.02\%         \\ \bottomrule[1pt]
\end{tabular}
}
\end{spacing}
\end{table}

\subsection{Source Data Collection}\label{sec: data_col}
To ensure \textit{ExpInstruct} covers real-world experimental scenarios, we choose video across three disciplines: science, healthcare and engineering, spanning 21 subjects for experiment commentary generation.
Specifically, we first collect these videos from YouTube\footnotemark\footnotetext{https://www.youtube.com} and Bilibili\footnotemark\footnotetext{https://www.bilibili.com} under Creative Commons licenses. Then, we recruit 10 college students from diverse scientific majors to review these videos. They filter out non-experimental and theory-only videos. After screening, we download the remaining videos with experiment titles, ranging from 360p to 720p, all at 30 FPS. Finally, we use WhisperX \cite{whisperx} for automatic speech recognition (ASR) to obtain timestamp-aligned video transcripts, annotated with second-level start and end timestamps.
\subsection{Automated Data Curation}\label{sec: data_cur}
To address issues in experiment commentary datasets (\eg, transcription errors, irrelevant content, and missing scientific principles and safety guidelines), we develop an automated data curation pipeline. This pipeline includes video transcript preprocessing, step-wise video clip segmentation, and principle-safety annotation. More details are provided in the Supplementary.

\vspace{0.5em}
\noindent\textbf{Video Transcript Preprocessing.}
Unlike generic video captioning, experiment commentary is rich in scientific terminology, presenting significant challenges for automatic speech recognition (ASR). ASR tools often mis-transcribe domain-specific terms, for instance, confusing ``titration'' with ``tight ration'' in chemistry experiments. To mitigate such errors, we refine ASR transcripts using GPT-4o \cite{gpt4}. We also translate multilingual transcripts into English to ensure linguistic consistency.

\vspace{0.5em}
\noindent\textbf{Step-wise Video Clip Segmentation.}
By leveraging the timestamp-aligned video transcripts, we use GPT-4o to segment the videos into clips that correspond to different procedural steps. This alignment enables accurate summarization and assignment of transcripts to each step (e.g., ``rinsing the burette with the standard solution'' and ``adding phenolphthalein to the conical flask''). It ensures synchronization between the video clip and its transcript (\ie, procedure).

\vspace{0.5em}
\noindent\textbf{Principle and Safety Guideline Annotation.}
We enhance commentary with annotations of scientific principles and safety guidelines. Our analysis of original procedures reveals that over 46\% of them require scientific principles and 35\% need safety guidance, yet less than 5\% and 3\% of original content include such information, respectively. To address this gap, we utilize GPT-4o to determine and generate necessary annotations. Specifically, we follow a structured prompt to extract relevant scientific principles and safety guidelines from experimental procedures. Finally, the procedure, along with associated scientific principles and safety guidelines, collectively form the commentary sequence $\{y_1, y_2, \dots, y_n\}$ for video clips.

\begin{figure}[t]
\centering
\begin{subfigure}{0.23\textwidth}
    \centering
    \includegraphics[width=\textwidth]{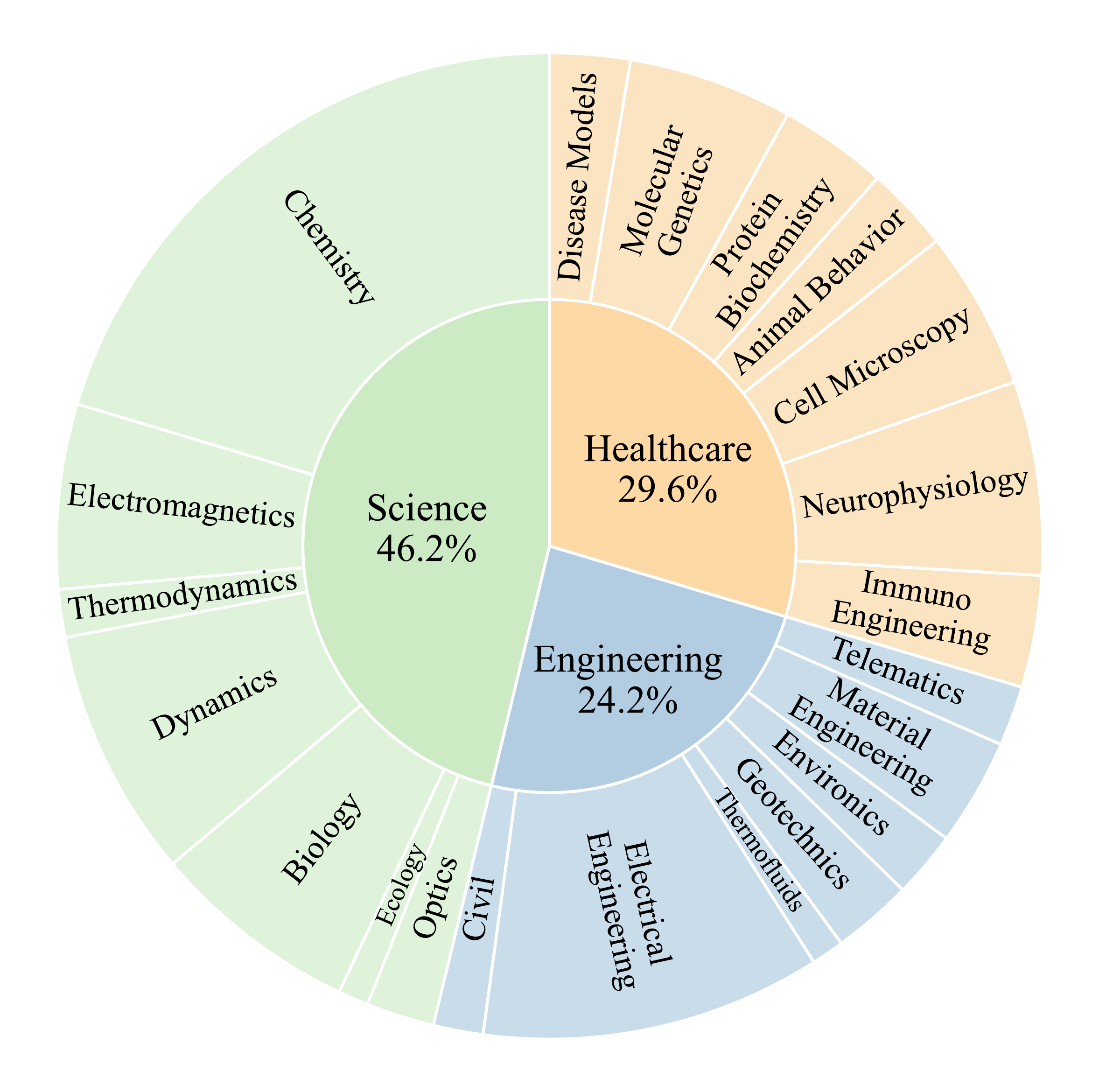}
    \caption{}
    \label{fig:pie}
\end{subfigure}
\hfill
\begin{subfigure}{0.23\textwidth}
    \centering
    \includegraphics[width=\textwidth]{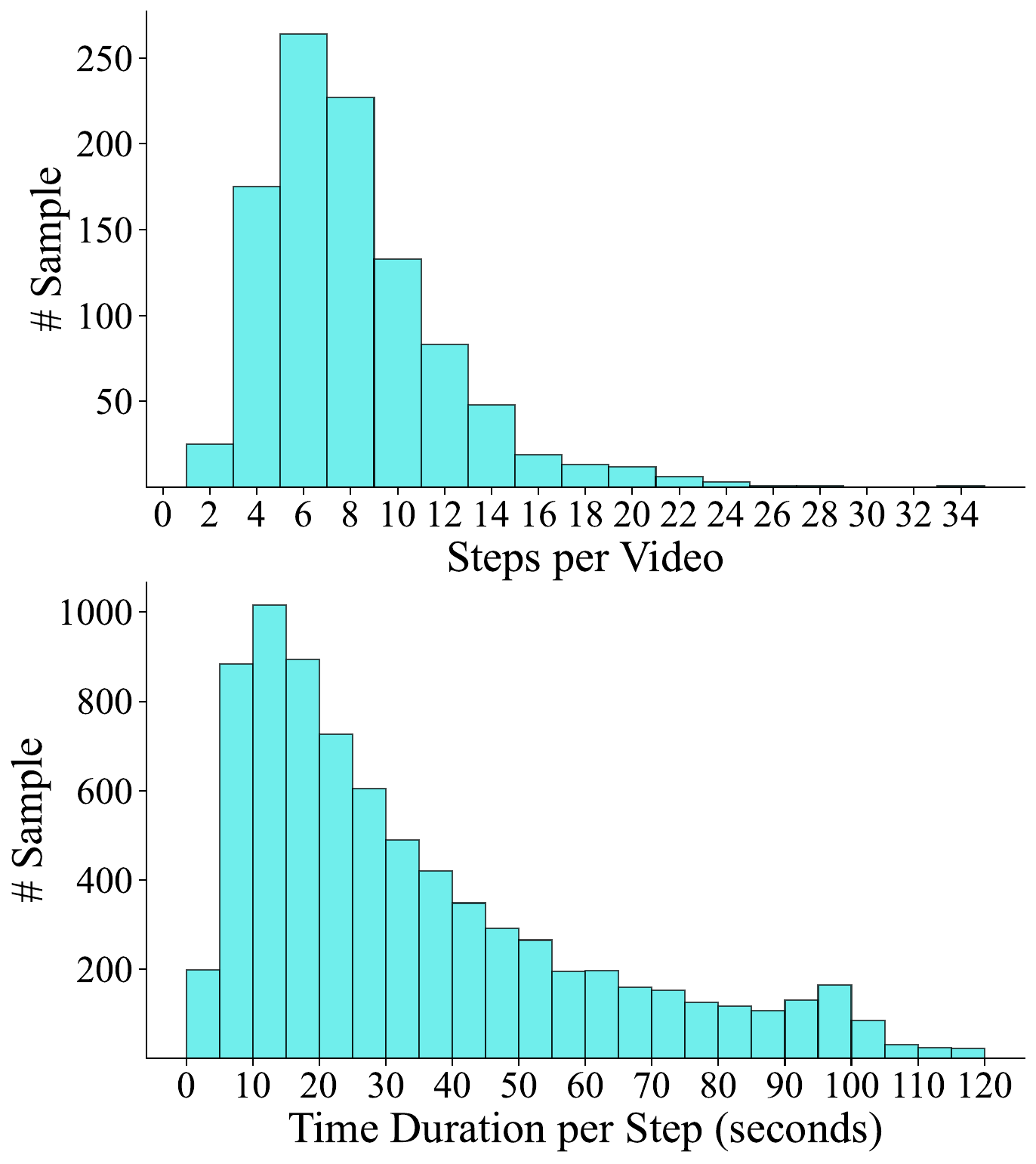}
    \caption{}
    \label{fig:merged}
\end{subfigure}
\caption{(a) Distribution of 21 subjects from 3 core disciplines in our \textit{ExpInstruct}. (b) Distribution of step numbers per video and the step duration.}
\label{fig: statis}
\end{figure}

\subsection{Data Quality Control}\label{sec: data_qua}

To ensure that \textit{ExpInstruct} maintains high-quality commentaries, avoiding annotation artifacts or irrelevant content, we implement a meticulous manual verification process. 10 college students from various scientific disciplines (\eg, chemistry, biomedical engineering) review data samples within their areas of expertise. Our verification process includes: \textbf{(i) Temporal Alignment Validation}: We verify the alignment between video clips and their commentaries, discarding or revising samples with misalignment. \textbf{(ii) Irrelevant Content Filtering}: We identify and remove video clips containing content unrelated to the experiment, such as general instructions, participant discussions, and non-experimental setups. \textbf{(iii) Commentary Accuracy Check}: We verify the accuracy of annotations (\ie, procedure, scientific principles, and safety guidelines). Inaccurate annotations (e.g., incorrect chemical equations) are either revised with minor edits or result in the removal of the entire video sample. We also eliminate overly cautious safety guidelines, such as ``Avoid spilling water to prevent accidents''.

\subsection{Dataset Analysis}\label{sec: data_ana}

We analyze the composition and characteristics of our \textit{ExpInstruct} dataset, with detailed statistics presented in Table \ref{tab: sta}. The dataset comprises 7,714 video clips from 1,011 experiment videos, split into 6,991 training samples and 723 testing samples. Each set (\ie, training or testing set) contains video clip-commentary pairs from distinct videos, enhancing the assessment of the model's generalization ability. \textbf{(i) Domain Distribution}:
\textit{ExpInstruct} spans 21 subjects across 3 core disciplines: science (3,565 samples, 46.2\%), healthcare (2,285 samples, 29.6\%), and engineering (1,864 samples, 24.2\%). The distribution of subjects and disciplines is illustrated in Fig. \ref{fig: statis} (a). \textbf{(ii) Length Analysis}:
Table \ref{tab: sta} and Fig. \ref{fig: statis} (b) present the word count of commentaries and the duration of video clips. Video clips range from 3 to 120 seconds, averaging 33 seconds, while commentaries average 50 words. The dataset encompasses 25,078 unique words, which significantly exceeds the vocabulary size of existing video captioning datasets, indicating a richer textual diversity with domain-specific scientific concepts. \textbf{(iii) Principle and Safety Guideline Statistics}:
Among the commentaries, 52.04\% include scientific principles and 36.42\% contain safety guidelines, underscoring the presence and importance of underlying principles and safety-relevant content beyond procedural descriptions. 
Our analysis
highlights \textit{ExpInstruct}'s comprehensive coverage of multi-disciplinary experimental scenarios for commentary generation.
\section{Proposed Method}
This section starts by formulating our commentary generation task (Sec. \ref{sec: pro_for}).
Next, we present our ExpStar model, which endows the LMM with adaptive knowledge retrieval and relevant passage judgment capabilities (Sec. \ref{sec: kb_gen}). 
Then, we design a rule-based reinforcement learning (RL) reward function to encourage ExpStar to generate safety guidelines aligned with experiments \ref{sec: safe_reward}). 
Finally, we elaborate on the inference strategy of our ExpStar (Sec. \ref{sec: inf}).

\begin{figure*}[]
  \centering
  \includegraphics[scale=0.483]{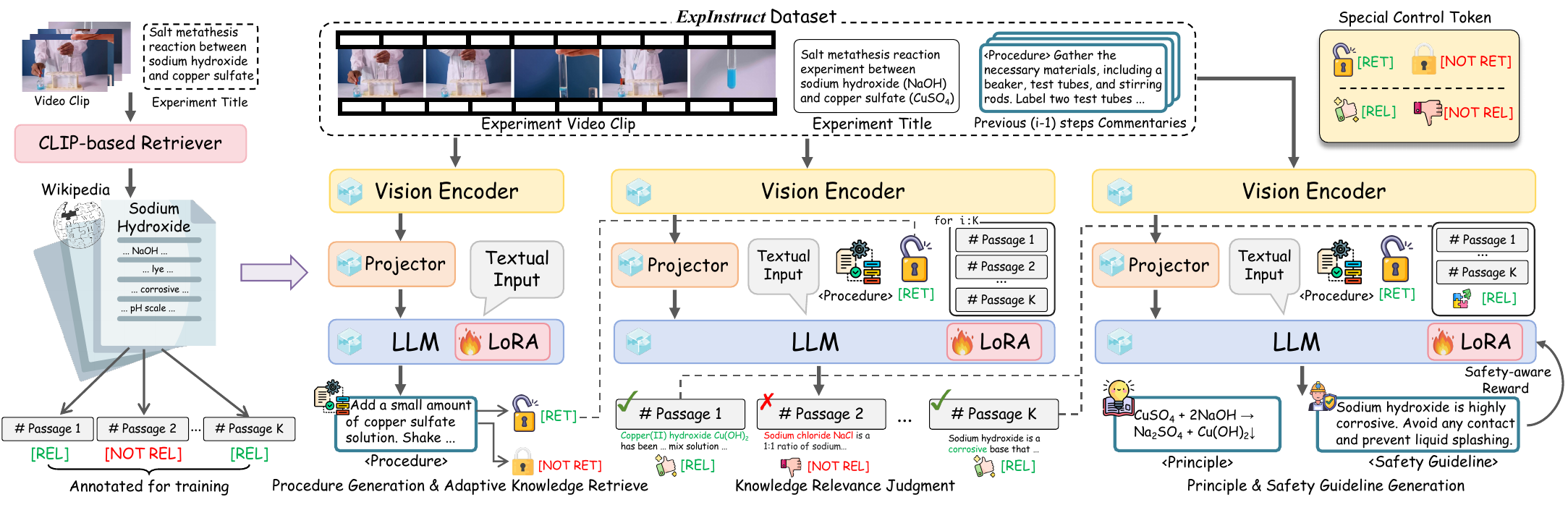}
  \caption{Overview of our proposed ExpStar model. It is built on the Qwen2.5-VL-7B architecture. Special control tokens are employed to guide the model's behavior: \texttt{<RET>} and \texttt{<NOT RET>} determine whether retrieval is necessary, while \texttt{<REL>} and \texttt{<NOT REL>} evaluate the relevance of retrieved passages. The right three LMMs share parameters.
  } 
  \label{fig: expstar_model}
\end{figure*}
\subsection{Problem Formulation}\label{sec: pro_for}
In our commentary generation task, given an experiment video $V = \{v_1, v_2, ..., v_n\}$ with $n$ steps and the experiment title $T$, our ExpStar model aims to generate a commentary $y_i$ for each video clip $v_i$, conditioned on contextual preceding commentary sequence $y_{<i}$. The commentary consists of basic procedures (\ie, an instructional description of the video clip), and optionally scientific principles and safety guidelines, depending on the context of the step. The objective of experiment commentary generation can be written as:
\begin{equation}
        y_i = \arg\max_{y_i} P(y_i \mid v_i, T, y_{<i}), \quad i = 1, 2, \dots, n, \\
\end{equation}
\begin{equation}
        y_i = [y_i^{\text{pro}};\ y_i^{\text{pri}};\ y_i^{\text{safe}}],
\end{equation}
where $y_i^{\text{pro}}$, $y_i^{\text{pri}}$, $y_i^{\text{safe}}$ represent procedure, principle and safety guideline, respectively.

Moreover, in our preliminary exploration, we observe that existing large multimodal models (LMMs) still struggle to 
produce accurate scientific principles and safety guidelines, as shown in Fig. \ref{fig: intro_case}.
It is primarily due to the lack of precise knowledge required to guide the generation of these elements, which undermines their applicability in realistic educational scenarios.
Therefore, our ExpStar determines whether additional information (\ie, scientific principles or safety guidelines) is required. If needed, it retrieves relevant external knowledge (\ie, textual passages) to guide the generation process.
Given that retrieved passages may introduce noise \cite{selfrag}, we further empower the LMM to assess the relevance of each passage during commentary generation.
The overview of our ExpStar model is illustrated in Fig. \ref{fig: expstar_model}.

\subsection{Knowledge-Augmented Generation}\label{sec: kb_gen}

\subsubsection{\textbf{Generation Protocol}}
Our ExpStar model begins by generating procedures based on the provided information (\ie, the video clip $v_i$, experiment title $T$, and preceding commentaries $y_{<i}$). When generating scientific principles and safety guidelines, existing LMMs often produce ambiguous content if they rely solely on visual and procedural information. By adaptively incorporating external knowledge retrieval, however, LMMs can generate more precise content for scientific principles and safety guidelines. Inspired by previous studies \cite{selfrag,mmvqa},  we augment the LMM's vocabulary with four additional special tokens, \ie, \texttt{<RET>}, \texttt{<NOT RET>}, \texttt{<REL>}, \texttt{<NOT REL>}. These tokens serve as control signals, enabling LMM to adaptively decide whether to retrieve passages (\texttt{<RET>} or \texttt{<NOT RET>}) and evaluate their relevance (\texttt{<REL>} or \texttt{<NOT REL>}) for generating principles or safety guidelines. Building upon this approach, we train our ExpStar model to develop two core capabilities: adaptive knowledge retrieval and knowledge relevance judgment.

\subsubsection{\textbf{Adaptive Knowledge Retrieval}}
As mentioned above, the generation of principles and safety guidelines benefits from external knowledge. Therefore, we trigger retrieval when generating content-related scientific principles or safety guidelines, using the special token \texttt{<RET>}. For others, we insert \texttt{<NOT RET>} instead:
\begin{equation}
    y_i =
    \begin{cases}
        \text{LMM}(v_i, T, y_{<i}, y_i^{\text{pro}}, \texttt{<RET>}, \mathbf{S}_o), & \exists (y_i^{\text{pri}} \vee y_i^{\text{safe}}),  \\
        \text{LMM}(v_i, T, y_{<i}, y_i^{\text{pro}}, \texttt{<NOT RET>}), & \text{otherwise},
    \end{cases}
\end{equation}
if the current step requires generating scientific principles or safety guidelines ($\exists (y_i^{\text{pri}} \vee y_i^{\text{safe}})$), the model triggers external knowledge retrieval by using the \texttt{<RET>} token and incorporating candidate passages \(\mathbf{S}_o\). Otherwise, the model does not perform external knowledge retrieval and generates the \texttt{<NOT RET>} token. 
For each experimental step that involves retrieval, we begin by utilizing a frozen CLIP-based retriever \cite{clip, evaclip, viclip} to encode both video frames and the title of the experiment into a unified embedding space, thereby generating a multimodal query. In parallel, textual passages from our knowledge base, which consists of intro paragraphs of Wikipedia articles \cite{selfrag}, are encoded using the same encoder to construct a search index. By calculating the cosine similarity between the multimodal query and every entry in this search index, we identify and select the top-$K$ most similar passages as our candidate passages \(\mathbf{S}_o\). Furthermore, we examine various configurations of retrievers and retrieval queries (Sec. \ref{sec: ablation_study}), encompassing different retriever models and query inputs.

\subsubsection{\textbf{Knowledge Relevance Judgment}}
While coarse-grained retrieval provides a preliminary set of candidate passages, they may contain irrelevant noise or marginally related information \cite{mmvqa}, which may mislead or dilute the generation quality. 
To address this, we enable the LMM to perform fine-grained relevance judgment, allowing it to selectively utilize passages that are truly useful for generation.
Specifically, for samples marked with the retrieval control token \texttt{<RET>}, we train ExpStar to emit binary relevance control tokens: \texttt{<REL>} and \texttt{<NOT REL>}, which indicate whether the specific given passage should be used during principle or safety guideline generation. These tokens are inserted after each retrieved passage $s_j$, where $s_j \in \mathbf{S}_o$ denotes one of the top-$K$ candidate passage:
\begin{equation}
    [v_i, T, y_{<i}, y_i^{\text{pro}}, \texttt{<RET>}, s_j, \texttt{<REL>}/\texttt{<NOT REL>}].
\end{equation}

Due to the absence of passage relevance annotations, we automatically construct binary relevance labels for the top-$K$ retrieved passages using GPT-4o. For each sample requiring retrieval, we prompt GPT-4o with the experiment title, the corresponding ground truth step-level commentary, and the top-$K$ retrieved passages. 
GPT-4o assigns a relevance score from 1 (completely irrelevant) to 5 (completely relevant) to each passage based on how well it supports the ground truth principle or safety guideline. Passages scoring \(\geq 3\) are considered \textit{relevant}, while those scoring \(\leq 2\) are deemed \textit{irrelevant}. 
This scoring scheme allows us to generate binary labels for training the relevance control tokens (\ie, \texttt{<REL>}, \texttt{<NOT REL>}). This automated supervision enables ExpStar to learn fine-grained selection, utilizing highly relevant passages to generate precise principles or safety guidelines.

\subsubsection{\textbf{Model Training}}
After retrieving and annotating relevance, we construct the following mixture of sequences to jointly train our ExpStar:
\begin{equation}
\begin{aligned}
&[v_i, T, y_i^{\text{pro}}, \texttt{<NOT RET>}], \\
&[v_i, T, y_i^{\text{pro}}, \texttt{<RET>}, s_j, \texttt{<REL>}, 
 (y_i^{\text{pri}} \vee y_i^{\text{safe}})], \\
&[v_i, T, y_i^{\text{pro}}, \texttt{<RET>}, \bar{s}_j, \texttt{<NOT REL>},  (y_i^{\text{pri}} \vee y_i^{\text{safe}})],
\end{aligned}
\label{eq: train}
\end{equation}
where $s_j$ and $\bar{s}_j$ are relevant and irrelevant passages, respectively. The notation \((y_i^{\text{pri}} \vee y_i^{\text{safe}})\) indicates that at least one of \(y_i^{\text{pri}}\) (scientific principle) or \(y_i^{\text{safe}}\) (safety guideline) is generated.
We then apply a token-level cross-entropy loss over all ground-truth tokens within the target commentary sequences (\ie, $y_i^{\text{pro}}$, $y_i^{\text{pri}}$, $y_i^{\text{safe}}$), excluding contributions from the video input $v_i$, experiment title $T$, and retrieved passages ($s_j$, $\bar{s}_j$). This ensures that the model is supervised only on the generation of relevant textual outputs, while the auxiliary inputs serve solely as conditioning context.
Supervised fine-tuning (SFT) with learnable lightweight LoRA \cite{lora} layers is performed for language modeling.

\subsection{Safety-aware Reward Function}\label{sec: safe_reward}
After SFT, ExpStar demonstrates promising capabilities in generating basic procedures and integrating knowledge. However, it tends to omit safety guidelines, a shortcoming that poses significant risks given the importance of accurate, context-specific safety instructions for conducting experiments safely.
Manual reviews reveal that approximately one-third of cases require safety guidelines, but only 20\% actually include them. To address this limitation, we implement a rule-based reward mechanism aimed at enhancing the coverage of experiment-specific safety guideline generation. Specifically, we adopt the direct preference optimization (DPO) algorithm \cite{dpo} to refine ExpStar by comparing preferred and non-preferred predictions. We select all training samples annotated with safety guidance and apply top-$p$ sampling ($p=0.9$) to generate $L$ candidate outputs for each sample. We then construct comparison pairs by evaluating the presence or absence of correct, experiment-specific safety guidelines, using the ground-truth annotation as a reference.
This approach improves both the coverage and quality of safety guideline generation.
Through this meticulous refinement, we strive to ensure that the generated experimental commentary incorporates safety reminders properly. This promotes educational objectives, providing students with thorough safety instructions for confident, secure, and responsible experiment execution.

\subsection{Inference Strategy}\label{sec: inf}
During inference, ExpStar generates commentary for each experiment step in a staged and controlled manner. Given the video clip $v_i$, experiment title $T$, and preceding commentaries $y_{<i}$, the model first predicts the basic procedure $y_i^{\text{pro}}$.
To determine whether additional knowledge is required for generating the scientific principle or safety guideline, ExpStar generates a control token indicating whether to retrieve external passages:
\begin{equation}
    t_i \sim \text{ExpStar}(v_i, T, y_{<i}, y_i^{\text{pro}}), \quad t_i \in \{\texttt{<RET>}, \texttt{<NOT RET>}\}.
\end{equation}

If \texttt{<NOT RET>} is predicted, the model terminates generation. 
Otherwise, \texttt{<RET>} prediction triggers to retrieve a top-$K$ set of passages $\mathbf{S}_o = \{s_1, ..., s_k\}$ using the video and title as a multimodal query. For each retrieved passage $s_j$, the model then decides its relevance by emitting a binary relevance token:
\begin{equation}
    r_j \sim \text{ExpStar}(v_i, T, y_{<i}, y_i^{\text{pro}}, \texttt{<RET>}, s_j),
\end{equation}
where $r_j \in \{\texttt{<REL>}, \texttt{<NOT REL>}\}$. We then gather the subset of passages $\mathbf{S}_{rel} \subseteq \mathbf{S}_o$ that are marked as relevant ($r_j = \texttt{<REL>}$). These passages are fed back into the model for final generation:
\begin{equation}
    [y_i^{\text{pri}} \vee y_i^{\text{safe}}] \sim \text{ExpStar}(v_i, T, y_{<i}, y_i^{\text{pro}}, \texttt{<RET>}, \mathbf{S}_{rel}).
\end{equation}

Notably, the inference process incorporates the previously generated commentary as input for the next step of inference.
The overall inference protocol enables ExpStar to adaptively decide whether external information is necessary, filter out irrelevant content, and generate contextually grounded and educationally informative scientific principles and safety guidelines beyond basic procedure.
\section{Experiments}

\begin{table*}[]
\centering
\begin{spacing}{1.}
\caption{\label{tab: main} Quantitative results on commentary generation of baselines and our ExpStar. \textbf{Bold}: the maximum value in the column.}
\resizebox{1.\textwidth}{!}{%
\begin{tabular}{lcccccccc}
\toprule[1pt]
\multicolumn{1}{l|}{\textbf{Models}}                                                            & \textbf{BLEU-1} & \textbf{BLEU-2} & \textbf{BLEU-3} & \textbf{BLEU-4} & \textbf{METEOR} & \textbf{ROUGE$_L$} & \textbf{CIDer} & \textbf{BERTScore} \\ \midrule
\multicolumn{9}{c}{\textit{Proprietary LMMs (Few-Shot)}}                                                                      \\ \midrule
\multicolumn{1}{l|}{GLM-4V-Plus \cite{glm4}}                                      & 9.64            & 4.68            & 1.70            & 0.52            & 22.12           & 10.91               & 7.68           & 48.54              \\
\multicolumn{1}{l|}{Gemini-2.0-Flash \cite{gemini}}                                                          & 13.75           & 5.51            & 2.59            & 0.92            & 26.92           & 15.64              & 10.93          & 49.46              \\
\multicolumn{1}{l|}{GPT-4o-mini \cite{gpt4}}                                                               & 10.73           & 3.92            & 1.65            & 0.55            & 23.98           & 10.18              & 5.32           & 47.42              \\
\multicolumn{1}{l|}{GPT-4o \cite{gpt4}}                                                                    & 15.10           & 5.65            & 2.43            & 0.71            & 26.07           & 15.47              & 13.68          & 52.97              \\ \midrule
\multicolumn{9}{c}{\textit{Open-sourced LMMs (Fine-tuning)}}                                                                                                                                                                                       \\ \midrule
\multicolumn{1}{l|}{Qwen2.5-VL-3B \cite{qwen2.5-vl}}                                                             & 32.36           & 14.09           & 7.12            & 2.61            & 28.74           & 26.10              & 24.36          & 60.67              \\
\multicolumn{1}{l|}{InternVL-2.5-4B \cite{internvl2.5}}                                                           & 31.27           & 13.62           & 6.91            & 2.55            & 27.38           & 25.84              & 21.10          & 59.77              \\
\multicolumn{1}{l|}{Qwen2-VL-7B \cite{qwen2-vl}}                                                               & 33.52           & 12.17           & 6.33            & 2.61            & 30.74           & 28.77              & 31.33          & 60.46              \\
\multicolumn{1}{l|}{Qwen2.5-VL-7B \cite{qwen2.5-vl}}                                                                                             & 34.98           & 16.08           & 8.73            & 3.93            & 30.91           & 26.68              & 29.90          & 60.85                \\
\multicolumn{1}{l|}{InternVL-2.5-8B \cite{internvl2.5}}                                                           & 36.82           & 16.66           & 8.55            & 3.42            & 27.51           & 27.04              & 23.68          & 60.49              \\
\multicolumn{1}{l|}{LLaVA-Next-Video-7B \cite{llavanext}}                                                       & 24.14           & 10.81           & 5.68            & 2.44            & 30.33           & 27.40              & 27.27          & 61.03              \\
\multicolumn{1}{l|}{VideoNarrator-7B \cite{videonar}}                                                             & 30.23           & 14.52           & 7.69            & 2.93            & 23.87           & 24.36              & 25.44          & 61.15              \\
\multicolumn{1}{l|}{VideoLLaMA2-7B \cite{videollama2}}                                                            & 30.07           & 13.56           & 7.16            & 2.76            & 26.72           & 23.07              & 16.70          & 59.83              \\
\multicolumn{1}{l|}{Video-LLaVA-7B \cite{videollava}}                                                               & 35.89           & 15.80           & 7.89            & 2.73            & 25.38           & 25.04              & 17.51          & 59.12              \\
\multicolumn{1}{l|}{CogVLM2-Video-8B \cite{cogvlm2}}                                                             & 30.07           & 14.58           & 8.08            & 3.50            & 24.38           & 25.24              & 33.49          & 59.89              \\
\midrule
\multicolumn{9}{c}{\textit{Ours}}                                                                                                                                                                                                                   \\ \midrule
\multicolumn{1}{l|}{ExpStar w/o KB}                                                            & 35.57           & 17.51           & 9.50            & 4.12            & 28.74           & 26.37              & 28.37         & 60.96              \\
\multicolumn{1}{l|}{ExpStar always w/ \texttt{<RET>}}                            & 37.34           & 18.21           & 9.82            & 4.09            & 29.38           & 25.78              & 25.56          & 60.53              \\
\multicolumn{1}{l|}{ExpStar w/o \texttt{<REL>} / \texttt{<NOT REL>}} & 43.26           & 21.21           & 11.51           & 5.01            & 28.51           & 28.17              & 38.84          & 62.57              \\
\multicolumn{1}{l|}{ExpStar w/o DPO}                                                           & 45.14           & 22.87           & 12.79           & 5.81            & 28.77           & 28.68              & 37.35          & 62.43              \\
\multicolumn{1}{l|}{ExpStar}                                                                   & \textbf{46.49}  & \textbf{23.92}  & \textbf{13.53}  & \textbf{6.37}   & \textbf{29.28}  & \textbf{28.89}     & \textbf{40.96} & \textbf{62.80}     \\

\bottomrule[1pt]
\end{tabular}
}
\end{spacing}
\end{table*}

\begin{table}[]
\centering
\caption{\label{tab: human}Human evaluation results. Each value is reported as mean/standard deviation. \textbf{Bold}: the best performance.}
\begin{spacing}{1.}
\resizebox{0.6\columnwidth}{!}{
\begin{tabular}{l|ccc}
\toprule[1pt]
\textbf{Model} & \textbf{Flu} & \textbf{Ins} & \textbf{Sci} \\
\midrule
GPT-4o         & 1.57  & 1.44  & 1.28 \\
Qwen2.5-VL-7B  & 1.22  & 1.06  & 0.82 \\
ExpStar        & \textbf{1.63} & \textbf{1.55} & \textbf{1.51} \\
\bottomrule[1pt]
\end{tabular}}
\end{spacing}
\end{table}
\subsection{Implementation Details}
All experiments are conducted with four NVIDIA A100-80GB GPUs.
Our ExpStar model adopts Qwen2.5-VL-7B \cite{qwen2.5-vl} as the backbone LMM. 
Our training pipeline consists of two phases: 
\textbf{(i) Supervised fine-tuning (SFT)}, we adopt EVA-CLIP-8B \cite{evaclip} as the default retriever to obtain the top-$K$ passages ($K=5$) based on the experiment title and video clip in our main experiment. Following previous studies \cite{selfrag,ency-vqa}, our knowledge base is a subset of Wikipedia, comprising the introductory paragraphs of approximately 6 million entries.
Additionally, we explore alternative retrieval modes, including variations in input queries, retrievers, and the value of $K$ in our ablation study (Sec. \ref{sec: ablation_study}).
To support the integration of retrieval-related control tokens, we modify the original vocabulary of ExpStar by replacing the final four reserved special tokens with our custom-designed ones.
We then perform SFT on ExpStar using the retrieval-augmented training sequences described in Eq. \ref{eq: train}. 
The training configuration includes a frame rate of 1.0 FPS, a learning rate of 1e-4, and a maximum sequence length of 4096 tokens.
\textbf{(ii) Reinforcement refinement}, we refine ExpStar using direct preference optimization (DPO) to enhance the coverage of experiment-specific safety guideline generation. The training process maintains the same configurations as the SFT phase but critically reduces the learning rate to 1e-6 to stabilize training. 
Both phases apply LoRA fine-tuning, with a LoRA rank of 8 and 3 training epochs.

\subsection{Evaluation Metric}
\subsubsection{\textbf{Automatic Evaluation Metrics}}
Following previous work ~\cite{matchtime,streaming,views,causal-vqg},
we adopt various popular metrics to evaluate the quality of generated commentaries. These metrics include BLEU-1 to BLEU-4 (B@1 to B@4) \cite{bleu}, ROUGE$_L$ (R$_L$) \cite{rouge}, METEOR (M) \cite{meteor}, CIDEr (Cr) \cite{cider}, and BERTScore (BS) \cite{bert-score}. 
All metric values are computed with the official implementations provided by the Hugging Face evaluation toolkit\footnotemark\footnotetext{https://github.com/huggingface/evaluate}.

\subsubsection{\textbf{Human Evaluation Criteria}}
To complement automatic metrics, we conduct human evaluation on 200 test samples. Five annotators with experience in experimental subjects rate the generated commentaries based on three criteria: 
fluency (\textbf{Flu}) measures grammatical correctness and readability. Instructional clarity (\textbf{Ins}) evaluates whether the procedural step is clearly and coherently described. Scientific appropriateness (\textbf{Sci}) assesses the accuracy and relevance of principle or safety-related content. Each is scored on a 2-point scale, and final scores are averaged across annotators.

\subsection{Baseline and Ablation Models}
\subsubsection{\textbf{Baseline Models}}
We evaluate our ExpStar model against 14 leading baselines, encompassing 4 proprietary and 12 open-source LMMs, as shown in Table \ref{tab: main}.
Specifically, we adopt a textual one-shot setting for proprietary LMMs, \ie, GPT-4o/4o-mini \cite{gpt4}, Gemini-2.0-Flash \cite{gemini}, and GLM-4V-Plus \cite{glm4}.
For open-source baselines, we consider LMMs with different parameter scales from 3B to 12B, including generic LMMs (\eg, Qwen2-VL \cite{qwen2-vl}) and video-specialized LMMs designed to capture spatio-temporal dynamics (\eg, VideoLLaMA2 \cite{videollama2}).
More details on the implementation of baselines are provided in the Supplementary.

\subsubsection{\textbf{Ablation Models}}
To evaluate the effectiveness of the proposed control tokens and reward function, we conduct the ablation study to compare the following variants of the inference pipeline: we analyze the effect of knowledge usage by comparing variants that disable retrieval (\textbf{w/o KB}), skip relevance filtering (\textbf{w/o \texttt{<REL>}}, \ie, two random passages from top-5 retrieval), or enforce retrieval for all steps (\textbf{always w/ \texttt{<RET>}}). We also assess the impact of removing safety-aware reinforcement refinement (\textbf{w/o DPO}). Moreover, we conduct experiments on different retrieval modes, including retriever backbones (\ie, CLIP \cite{clip}, EVA-CLIP-8B ~\cite{evaclip} and ViCLIP ~\cite{viclip}), queries, and top-$K$ values.

\subsection{Performance Comparison}
Table \ref{tab: main} presents a comprehensive comparison between our ExpStar model and 14 leading LMMs. We have the following key findings: \textbf{(i) Limited effectiveness of proprietary LMMs.}
Despite their scale and general capabilities, proprietary models such as GPT-4o perform poorly on automatic language generation metrics. Upon closer inspection, we find that their generated commentaries are often overly verbose, frequently appending scientific principles and safety guidelines to nearly every procedural step regardless of necessity. 
It diverges from realistic experiment scenarios, where time-constrained experiment videos require concise and context-sensitive commentary on each step. 
\textbf{(ii) Direct supervised fine-tuning alone on open-sourced LMMs still fall short.}
While open-source LMMs show performance gains through end-to-end supervised fine-tuning and spatial-temporal mechanisms, they still struggle to generate scientifically grounded and contextually appropriate scientific principles and safety guidelines within the commentaries. 
Specifically, our 7B ExpStar model significantly outperforms the larger InternVL-2.5-8B model by +2.95 and +17.28 on BLEU-4 and CIDEr, respectively. 
It also consistently surpasses video-centric LMMs (\ie, LLaVA-Next-Video, VideoNarrator, VideoLLaMA2, Video-LLaVA and CogVLM-Video) across all metrics.
This underscores that simply scaling up model size and incorporating spatial-temporal features are not sufficient; effective integration of external knowledge is critical for the LMM to produce accurate experiment commentary when principles or safety guidelines are required.
\textbf{(iii) ExpStar shows stronger applicability to realistic experimental settings.}
In Table \ref{tab: human}, ExpStar achieves the highest scores across all criteria, including fluency, instructional clarity, and scientific appropriateness. Although our model is built upon the same backbone as Qwen2.5-VL-7B, it significantly outperforms it in human ratings, highlighting the effectiveness of our adaptive knowledge-augmented design and post direct preference optimization(DPO) training for commentary generation.

\begin{table}[]
\centering
\begin{spacing}{1.}
\caption{\label{tab: ret} Performance of different retrievers and retrieval queries. V, T and P represent the video clip, experiment title and basic procedure, respectively. \textbf{Bold}: best values.}
\resizebox{1.0\columnwidth}{!}{%
\begin{tabular}{lc|ccccc}
\toprule[1pt]
\textbf{Retrievers} & \textbf{Queries} & \textbf{B@4} & \textbf{M} & \textbf{R$_L$} & \textbf{Cr} & \textbf{BS} \\ \midrule
\multirow{3}{*}{CLIP \cite{clip}} 
            & V                     & 5.77         & 28.16      & 27.68         & 35.08 & 62.04       \\
             & V+T             & 5.89         & 28.25      & 28.20          & 36.51       & 62.26     \\
             & V+T+P           & 5.80        & 28.05      & 28.41         & 35.82       & 62.17       \\ \midrule
\multirow{3}{*}{EVA-CLIP \cite{evaclip}} 
            & V                     & 5.87         & 28.76      & 28.68         & \textbf{39.08} & 62.38       \\
             & V+T             & \textbf{6.15} & \textbf{29.15} & \textbf{29.03} & 38.99       & \textbf{62.94} \\
             & V+T+P           & 5.93         & 28.75      & 28.61         & 38.82       & 62.60       \\ \midrule
\multirow{3}{*}{ViCLIP \cite{viclip}} 
            & V                     & 4.42         & 28.37     & 28.09         & 38.13 & 61.96       \\
             & V+T             & 4.59         & 28.42      & 28.13         & 38.41       & 62.11 \\
             & V+T+P           & 4.48         & 28.15      & 27.92         & 37.96       & 62.19       \\
\bottomrule[1pt]
\end{tabular}%
}
\end{spacing}
\end{table}

\begin{table}[]
\centering
\begin{spacing}{1.}
\caption{\label{tab: topk} Performance comparison across different model sizes with top-$K$ retrieval. \textbf{Bold}: best values.}
\resizebox{1.0\columnwidth}{!}{%
\begin{tabular}{cc|cccccc}
\toprule[1pt]
\textbf{ExpStar} & \textbf{$K$} & \textbf{B@1} & \textbf{B@4} & \textbf{M} & \textbf{R$_L$} & \textbf{Cr} & \textbf{BS} \\ \midrule
\multirow{5}{*}{\footnotesize Qwen2.5-VL-7B} 
 & 0 & 35.57                      & 4.12            & 28.74           & 26.37              & 28.37         & 60.96  \\ 
 & 1 & 45.92 & 5.93 & 29.09 & 28.84 & 39.51 & 62.86 \\ 
 & 3 & \textbf{46.58} & 6.15 & 29.15 & \textbf{29.03} & 38.99 & \textbf{62.94} \\
 & 5 & 46.49 & \textbf{6.37} & \textbf{29.28} & 28.89 & \textbf{40.96} & 62.80 \\
 & 8 & 45.41 & 5.74 & 28.90 & 28.50 & 38.36 & 62.77 \\ \midrule
 
\multirow{5}{*}{\footnotesize Qwen2.5-VL-3B} 
 & 0 &36.51 & 3.38 &26.11 & 24.91 & 25.05 &58.68 \\ 
 & 1 & 38.49 & \textbf{3.86} & 25.11 & 24.26 & 25.63 & 59.90 \\ 
 & 3 & \textbf{39.35} & 3.32 & 25.30 & 24.62 & 25.52 & 59.82 \\
 & 5 & 39.05 & 3.72 & \textbf{26.71} & \textbf{25.37} & \textbf{32.27} & \textbf{60.54} \\
 & 8 & 38.77 & 3.43 & 25.88 & 24.87 & 27.33 & 60.15 \\ 
\bottomrule[1pt]
\end{tabular}%
}
\end{spacing}
\end{table}

\subsection{Ablation Study}\label{sec: ablation_study}
The results of conducted ablation experiments are shown in Table \ref{tab: main}, Table \ref{tab: ret}, and Table \ref{tab: topk}. 
Our key findings are as follows: \textbf{(i) External knowledge is the key to performance gains.}
Without knowledge retrieval (w/o KB), ExpStar's CIDEr score drops from 40.96 to 28.37 and BLEU-4 from 6.37 to 4.12 in Table \ref{tab: main}, highlighting the indispensable role of knowledge grounding in commentary generation, especially for explaining scientific principles and safety considerations.
\begin{figure*}[]
  \centering
  \includegraphics[scale=0.56]{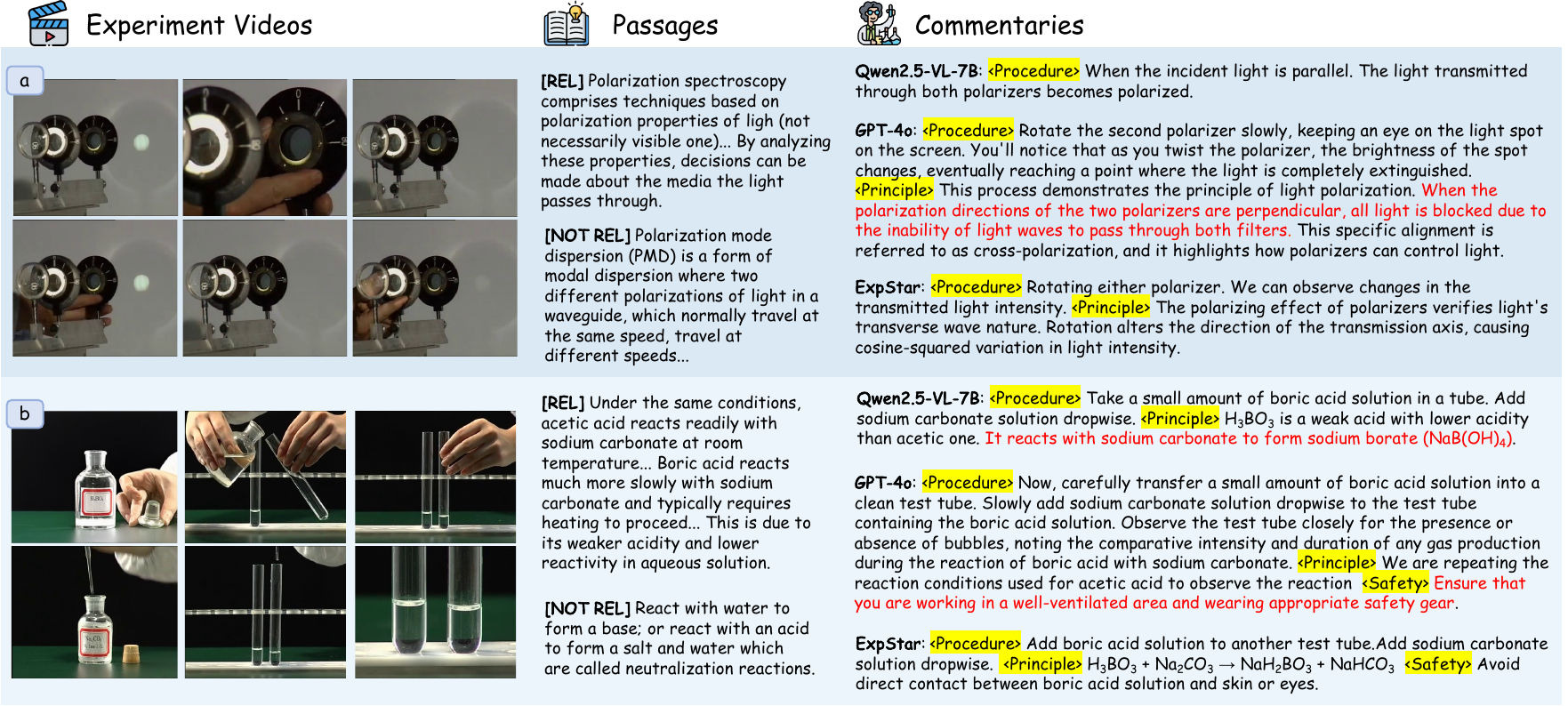}
  \caption{Qualitative results of Qwen-2.5-VL-7B, GPT-4o and our ExpStar. The \textcolor[RGB]{255,0,0}{red} text denotes inappropriate generation.
  } 
  \label{fig: case_study}
\end{figure*}
\textbf{(ii) Blindly retrieving knowledge for all steps introduces noise.}
The variant that always applies the \texttt{<RET>} token (\ie, ExpStar always w/ \texttt{<RET>}) regardless of step necessity underperforms the completed ExpStar model, particularly on CIDEr (25.56 vs. 40.96).
This confirms the effectiveness of our adaptive retrieval decision mechanism for suppressing irrelevant knowledge usage.
\textbf{(iii) Relevance control tokens enhance knowledge utilization.} Skipping the use of \texttt{<REL>} / \texttt{<NOT REL>} and instead randomly selecting passages from top-5 results leads to a noticeable drop in generation quality. 
BLEU-4 decreases by 1.36 and CIDEr by 2.12, demonstrating the value of fine-grained relevance supervision in filtering noise from retrieved passages.
\textbf{(iv) Safety-aware optimization enhances instructional precision.}
While ExpStar already captures procedural description and scientific principles after supervised fine-tuning (SFT), removing safety-specific reward optimization (\textbf{w/o DPO}) results in noticeably less precise and less frequent generation of safety commentary.
The precision of safety commentary generation, defined as the alignment between model predictions and ground truth regarding the generation or omission of safety guidelines, declined from 87.45\% to 62.30\%. Concurrently, the frequency of safety guidelines predictions decreases from 34.27\% to 20.93\%.
This gap confirms that our rule-based safety-aware reward is crucial for encouraging the model to generate safety-relevant content when necessary — an essential aspect of scientific experiment instruction often overlooked by LMMs during direct SFT.
\textbf{(v) Impact of retrieval combinations on performance differentiation.}
As shown in Table \ref{tab: ret}, among various retriever-query combinations, EVA-CLIP with video and title (\ie, V+T) yields the best overall performance. Incorporating procedure (V+T+P) slightly degrades results, likely due to increased redundancy in the query, which narrows the retrieval scope of knowledge.
Table \ref{tab: topk} further shows that setting $K=5$ achieves the best trade-off between informativeness and noise.
Additionally, we observe consistent gains from our retrieval mode for different parameter sizes of the LMMs (\ie, 3B and 7B variants of ExpStar), as shown in Table \ref{tab: topk}.
Moreover, performance significantly improves when scaling our backbone (\ie, Qwen2.5-VL) from 3B to 7B parameters. 

\subsection{Qualitative Comparisons}
Fig. \ref{fig: case_study} provides the qualitative comparisons on our \textit{ExpInstruct} dataset. ExpStar consistently produces scientifically grounded and pedagogically appropriate (\ie, concise) commentary by leveraging retrieved passages. Importantly, its ability to assess the relevance of each passage (\ie, \texttt{<REL>} or \texttt{<NOT REL>}) ensures that only contextually appropriate knowledge for generation. More cases from different disciplines are provided in Supplementary.

\section{Conclusion}
In this paper, we present ExpStar, the first model designed to generate commentaries for multi-discipline scientific experiment videos automatically. 
To facilitate this task, we construct the \textit{ExpInstruct} dataset, which includes over 7\textit{K} step-level commentary annotations covering 21 scientific subjects. These annotations encompass basic procedures, underlying scientific principles, and safety guidelines. 
ExpStar integrates adaptive retrieval mechanisms and fine-grained relevance control over external knowledge, further enhanced by safety-aware reward optimization. These innovations enable ExpStar to produce accurate and contextually relevant scientific principles and safety guidelines.
Extensive experimental evaluations demonstrate that ExpStar outperforms existing large multimodal models (LMMs), offering a practical and effective solution for AI-driven assistance in scientific experiment instruction.
\definecolor{boxorange}{HTML}{FFA500} 
\definecolor{boxblue}{HTML}{0000FF}   
\definecolor{boxred}{HTML}{FF0000}    
\newcommand{\algokeyword}[1]{\texttt{#1}}
\newcommand{\orangebox}[1]{%
  \colorbox{boxorange!30}{\algokeyword{#1}}
}
\newcommand{\bluebox}[1]{%
  \textcolor{boxblue}{\fbox{\algokeyword{#1}}}
}
\newcommand{\redbox}[1]{%
  \textcolor{boxred}{\fbox{#1}}
}

\bibliographystyle{ACM-Reference-Format}
\balance
\bibliography{main}

\newpage
\appendix
In ~\cref{sec: prompt}, we present the prompts used for dataset annotation, detailing the instructions and formatting designed to elicit high-quality responses from annotators.  
In \cref{sec: Pseudocode}, we describe the training and inference pseudocode of our ExpStar.
In \cref{sec: Retrieval}, we explain the coarse-grained video-text retrieval framework with different retrievers and retrieval
queries.
Moreover, \cref{sec: baseline} offers configuration details and training setups that ensure fair evaluation. 
Finally, we provide additional qualitative cases to supplement the main experiments in ~\cref{sec: case}.

\section{\textit{ExpInstruct} Dataset Construction Detials} \label{sec: prompt}

In this section, we first provide the prompts for AI-assisted annotation.
For manual verification, we 
elaborate on the demographics of our human annotators.

\begin{algorithm*}[h]
\vspace{2mm}
\caption{\textbf{Algorithm 1 ExpStar Training}}
\label{al: train}
\vspace{2mm}
\begin{algorithmic}[1]
\REQUIRE Retriever $\mathcal{R}$,  Knowledge base document $\mathcal{D}$
\STATE \textbf{Input:} baseline training dataset $\mathcal{P}$ = $\{X, Y\}$, Generator LM $\mathcal{M}_{origin}$, \textbf{Output:} Generator LM $\mathcal{M}_{finetune}$
\FOR{$ (x, y) \in (X, Y) $}
\IF{$y^{\text{pri}}$ \textbf{or} $y^{\text{safe}}$ in $y$}
\STATE Retrieve top $k$ relevant passage $\{d_1, d_2, \dots, d_k\}$ in knowledge base document $\mathcal{D}$ using $\mathcal{R}$ 
\FOR{$i=1$ to $k$}
\STATE Add \texttt{<RET>} after $y^{\text{pro}}$ 
\STATE Prompt GPT-4o to predict \texttt{<REL>}/\texttt{<NOT REL>}
    \IF {\texttt{<REL>}}
        \STATE Add $d_i$ and \texttt{<REL>} after \texttt{<RET>} to form $ [x, y^\text{pro}, \texttt{<RET>}, d_i, \texttt{<REL>}, (y_i^{\text{pri}} \vee y_i^{\text{safe}})] \in \mathcal{P'} $
    \ELSIF{\texttt{<NOT REL>}}
        \STATE Add $d_i$ and \texttt{<NOT REL>} after \texttt{<RET>} to form $ [x, y^\text{pro}, \texttt{<RET>}, d_i, \texttt{<NOT REL>}, (y_i^{\text{pri}} \vee y_i^{\text{safe}})] \in \mathcal{P'} $
    \ENDIF 
\ENDFOR
\ELSE
\STATE Add \texttt{<NOT RET>} after $y^{\text{pro}}$ to form $ [x, y^\text{pro}, \texttt{<NOT RET>}] \in \mathcal{P'} $
\ENDIF
\ENDFOR
\STATE Obtain ExpStar training dataset $\mathcal{P'}$
\STATE Update $\mathcal{M}_{origin}$ on $\mathcal{P'}$ with next token prediction loss to get $\mathcal{M}_{finetune}$
\end{algorithmic}
\end{algorithm*}

\begin{algorithm*}[h]
\vspace{2mm}
\caption{\textbf{Algorithm 2 ExpStar Inference}}
\vspace{2mm}
\label{al: inference}
\begin{algorithmic}[1]
\REQUIRE Generator LM $\mathcal{M}$, Retriever $\mathcal{R}$,  Knowledge base document $\mathcal{D}$
\STATE \textbf{Input:} input prompt $x$ and preceding generation $y_{<t}$, \textbf{Output:} next output segment $y_t$
\STATE $\mathcal{M}$ predicts \texttt{<RET>}/\texttt{<NOT RET>} given $(x, y_{<t})$
\IF {\texttt{<RET>}} 
\STATE Initialize an empty relevant document set $\mathcal{S}_{rel}$
\STATE Retrieve top $k$ passage $\{d_1, d_2, \dots, d_k\}$ in knowledge base document $\mathcal{D}$ using $\mathcal{R}$
\FOR{$i=1$ to $k$}
\STATE $\mathcal{M}$ predicts \texttt{<REL>}/\texttt{<NOT REL>} for $d_i$ given $(x, y_{<t})$ 
    \IF {\texttt{<REL>}} 
        \STATE Add $d_i$ to $\mathcal{S}_{rel}$
    \ELSIF{\texttt{<NOT REL>}}
        \STATE continue
    \ENDIF 
\ENDFOR
\STATE $\mathcal{M}$ predicts $y_t$ given $(x, y_{<t}, \mathcal{S}_{rel})$
\ELSIF{\texttt{<NOT RET>}}
\STATE $y_t$ = \texttt{<EOS>}
\ENDIF 
\end{algorithmic} 
\end{algorithm*}

\subsection{Prompt for Dataset Annotation}
Table \ref{tab:prompt_1} presents the prompt utilized for correcting errors in ASR-generated transcriptions. Table \ref{tab:prompt_2} elaborates on the prompt created for step allocation and summarization within procedural tasks. Table \ref{tab:prompt_3} describes the prompt designed to assist annotators in labeling scientific principles and safety guidelines. Finally, Table \ref{tab:prompt_4} details the prompt developed for evaluating the relevance between ground truth principles or safety guidelines and their corresponding retrieval passages.

\begin{table*}[]
    \centering
    \small
    \begin{spacing}{1.05}
    \caption{\label{tab:prompt_1} The prompt for Correcting Errors in ASR-Generated Text. \texttt{[Subject]} indicates the subject name when processing experimental videos from different disciplines.}
    \resizebox{\textwidth}{!}{
    \begin{tabular}{p{\linewidth}}
    \toprule[1pt]
You are a \texttt{[Subject]} advisor. Please correct any spelling errors, especially homophones used incorrectly, as well as mistakes \texttt{[Subject]} formatting or terminology in the experimental procedure text. Directly update the "text" field with the corrected version.\\
Note:\\
-If there are no errors, do not make any changes. Still, return all items in JSON format (including unchanged content). \\
-Do not include any extra explanation—only return the updated JSON.

\texttt{Template:}\\
\texttt{\{}\\
\texttt{\ \ \ \ "id": "",}\\
\texttt{\ \ \ \ "startTime": "",}\\
\texttt{\ \ \ \ "endTime": "",}\\
\texttt{\ \ \ \ "text": ""}\\
\texttt{\}}\\
    \midrule[1pt]
    \end{tabular}
    }
    \end{spacing}
\end{table*}

\begin{table*}[]
    \centering
    \small
    \begin{spacing}{1.05}
    \caption{\label{tab:prompt_2} The prompt for Step Summarization. \texttt{[Subject]} indicates the subject name when summarizing the steps of experiments from different academic disciplines.}
    \resizebox{\textwidth}{!}{
    \begin{tabular}{p{\linewidth}}
    \toprule[1pt]
You are an experiment instructor for the subject of \texttt{[Subject]}. Based on the ASR JSON text of the experiment audio, please: \\
1. Generate the "summary" field (experiment summary); \\
2. Summarize the experiment procedure into the "steps" field, with numbered steps starting from 1; \\
3. Generate a new field "procedure" (a summary description of each step); \\
4. Generate the "ASR\_id" field, which must strictly follow the order of the ASR fragments. Each step's ASR fragments should consist of a continuous sequence of positive integer numbers (e.g., [2,3,4,5]), without any jumps or regressions (e.g., [2,3,7,5] is not allowed); \\
5. Each ASR fragment must only appear in one step and cannot be repeated in different steps. \\
Note: \\
- Only return the JSON result, no additional explanation is required. \\
- Ensure that the "ASR\_id" only contains continuous positive integer numbers, without any skips or backtracking. \\

\texttt{Template:} \\
\texttt{\{} \\
\texttt{\ \ \ \ "summary":"",} \\
\texttt{\ \ \ \ "steps": [} \\
\texttt{\ \ \ \ \ \ \ \ \{}\\
\texttt{\ \ \ \ \ \ \ \ \ \ \ \  "step":1,} \\
\texttt{\ \ \ \ \ \ \ \ \ \ \ \  "procedure":"",} \\
\texttt{\ \ \ \ \ \ \ \ \ \ \ \  "ASR\_id":[x, x],} \\
\texttt{\ \ \ \ \ \ \ \ \},}\\
\texttt{\ \ \ \ \ \ \ \ \{}\\
\texttt{\ \ \ \ \ \ \ \ \ \ \ \  "step":2,} \\
\texttt{\ \ \ \ \ \ \ \ \ \ \ \  "procedure":"",} \\
\texttt{\ \ \ \ \ \ \ \ \ \ \ \  "ASR\_id":[x, x]} \\
\texttt{\ \ \ \ \ \ \ \ \},}\\
\texttt{\ \ \ \ \ \ \ \  ......}\\
\texttt{\ \ \ \ \ ]} \\
\texttt{\}} \\
    \midrule[1pt]
    \end{tabular}
    }
    \end{spacing}
\end{table*}

\begin{table*}[]
    \centering
    \small
    \begin{spacing}{1.05}
    \caption{\label{tab:prompt_3} The prompt for Scientific Principles and Safety Guidelines Annotation. \texttt{[Subject]} indicates the subject name when processing experimental texts from different disciplines.}
    \resizebox{\textwidth}{!}{
    \begin{tabular}{p{\linewidth}}
    \toprule[1pt]
You are a \texttt{[Subject]} experiment instructor. \\
Please generate a new field \texttt{"safety"} for each procedure sentence to indicate relevant safety concerns. Avoid redundant warnings for routine or low-risk operations, and do not overemphasize basic or common-sense precautions. \\
Also, generate a new field \texttt{"principle"} to identify relevant \texttt{[Subject]} equations or core scientific principles involved in each operation. \\
Return the result in JSON format only (no additional explanation). \\
Note: \\
- Not every sentence involves safety issues or scientific principles. If a sentence does not relate to either, do not add new fields—keep the original structure unchanged. \\
- Do not modify existing fields—only add new ones where applicable. \\

\texttt{Template:} \\
\texttt{\{} \\
\texttt{\ \ \ \ "id": "",} \\
\texttt{\ \ \ \ "startTime": "",} \\
\texttt{\ \ \ \ "endTime": "",} \\
\texttt{\ \ \ \ "text": "",} \\
\texttt{\ \ \ \ "safety": "", \ \ \# Optional, only if relevant} \\
\texttt{\ \ \ \ "principle": "" \ \ \# Optional, only if relevant} \\
\texttt{\}} \\
    \midrule[1pt]
    \end{tabular}
    }
    \end{spacing}
\end{table*}
\begin{table*}[]
    \centering
    \small
    \begin{spacing}{1.05}
    \caption{\label{tab:prompt_4} The Prompt for Evaluating Relevance Between ground truth principle or safety guideline and retrieval passage.}
    \resizebox{\textwidth}{!}{
    \begin{tabular}{p{\linewidth}}
    \toprule[1pt]
You are a professional text relevance evaluation assistant. Your task is to evaluate the relevance between a query and a document. \\
Please provide a score from 1 to 5 based on the degree of relevance between the query and the document, where: \\
\quad 1: Completely irrelevant \\
\quad 2: Slightly relevant \\
\quad 3: Moderately relevant \\
\quad 4: Highly relevant \\
\quad 5: Completely relevant \\
The input will be provided as follows: \\
\texttt{Query}: The ground truth principle or safety guideline. \\
\texttt{Document}: The retrieval passage from the knowledge base. \\
Please return only an integer score between 1 and 5 without any additional explanation. \\
    \midrule[1pt]
    \end{tabular}
    }
    \end{spacing}
\end{table*}

\subsection{Annotator Demographics}
Specifically, we recruit a total of ten annotators with diverse academic backgrounds, including chemistry, biology, physics, and medicine. The team consists of three female master's students, five male master's students, and two male PhD students. All annotators possess strong domain knowledge and relevant academic training, enabling them to comprehend and assess task-related content.

\section{Training and Inference Overview of ExpStar} \label{sec: Pseudocode}
Algorithm \ref{al: train} and Algorithm \ref{al: inference} demonstrate the overview of ExpStar at training and inference, respectively.

\vspace{18pt} 
\section{Coarse-Grained Video-Text Retrieval Mechanism} \label{sec: Retrieval}
Our methodology leverages multimodal latent representations derived from both visual and textual modalities as anchors for retrieval. For video processing, we first temporally sample the input video V at fixed intervals to obtain a sequence of key frames, which are then encoded along with associated textual information (\eg experiment title T and basic procedure P) to construct the initial candidate set $\mathcal{S}_0$. The retrieval pipeline employs three CLIP-based encoders for feature extraction: 
\begin{itemize}[leftmargin=*]
\item \texttt{OpenAI/CLIP} provides robust baseline cross-modal alignment. \item \texttt{EVA-CLIP-8B} is an enhanced large-scale variant offering superior representation capabilities. 
\item \texttt{ViCLIP} is optimized explicitly for temporal video understanding. These encoders generate dense feature representations $\mathbf{z}_i \in \mathbb{R}^d$ (where $d$ denotes the embedding dimensionality), which collectively form our search index $\mathcal{Z} = \{\mathbf{z}_i\}_{i=1}^N$. 
\end{itemize}

To compute query-document similarity, we implement three multimodal fusion strategies operating on the embedding space:  
\begin{itemize}[leftmargin=*]
\item \textit{Video-Title Fusion} (V+T): it combines visual features $\mathbf{q}_v$ (derived from frame-wise CLIP feature averaging) with title embeddings $\mathbf{q}_t$ through weighted summation ($0.7\mathbf{q}_v + 0.3\mathbf{q}_t$); \item \textit{Video-Title-Procedure Fusion} (V+T+P): it extends the textual information by incorporating both title and procedures embeddings $\mathbf{q}_{tn}$ ($0.7\mathbf{q}_v + 0.3\mathbf{q}_{tn}$).
\item \textit{Video-Only Retrieval} (V): it relies exclusively on visual embeddings $\mathbf{q}_v$. All similarity computations employ the standard cosine similarity metric. The retrieval mechanism selects the top-$k$ most relevant passages based on the similarity scores, with aggregated results constituting the final candidate set $\mathcal{S}_0$.
\end{itemize}
This mechanism demonstrates significant flexibility in accommodating diverse query configurations while maintaining robust retrieval performance across various multimodal combinations.

\section{Baseline Details}  \label{sec: baseline}
\subsection{Configuration for Proprietary LMMs}
Table \ref{tab:prompt_5} details the custom prompt architecture used with proprietary large multimodal models in our experiments. For the baseline setup with closed-source models, we adopt a textual one-shot setting, where a randomly selected training sample is provided as context without visual input. The input videos are uniformly sampled into 32 frames.

\begin{table*}[]
    \centering
    \small
    \begin{spacing}{1.05}
    \caption{\label{tab:prompt_5} The prompt for Generating procedure with Scientific and Safety Annotations in Biology Experiments.}
    \resizebox{\textwidth}{!}{
    \begin{tabular}{p{\linewidth}}
    \toprule[1pt]
You are an expert narrator for biology experiment videos, helping students better understand biological concepts, experimental processes, and observational results. \\
Please always use \texttt{<Procedure>} tags to mark the procedure content. When necessary, also use the following tags: \\
1. \texttt{<Principle>}: to annotate explanations of scientific principles or theories. \\
2. \texttt{<Safety>}: to annotate safety precautions or warnings. \\

Please generate a detailed procedure for the current step of the biology experiment titled \texttt{"Detecting Sugars, Fats and Proteins in Biological Tissues"}. \\
The images provided are equidistant samples taken from a video. Keep your responses as concise as possible. \\

\texttt{Example Response:} \\
\texttt{<Procedure>} Add 1.5g of copper(II) sulfate pentahydrate crystals to test tube No. 3 \\
\texttt{<Principle>} CuSO$_4$$\cdot$5H$_2$O(s) $\rightarrow$ Cu$^{2+}$(aq) + SO$_4^{2-}$(aq) + 5H$_2$O(l) \\
    \midrule[1pt]
    \end{tabular}
    }
    \end{spacing}
\end{table*}

Due to the absence of video input capabilities in proprietary large multimodal models, we process video clips by extracting frames at 1 FPS. A maximum of ten equally spaced frames is selected as inputs for the model. If the total number of sampled frames is fewer than ten, all available frames are utilized.

\subsection{Deployment and Inference Details}
All experiments are carried out using four NVIDIA A100 GPUs with 80GB memory each.
For open-source models, we utilize the vLLM framework\footnote{https://github.com/vllm-project/vllm} for local deployment, relying on their official checkpoints for model serving.

\section{More Cases} \label{sec: case}

\begin{figure*}[h]
    \centering
    \includegraphics[width=1\linewidth]{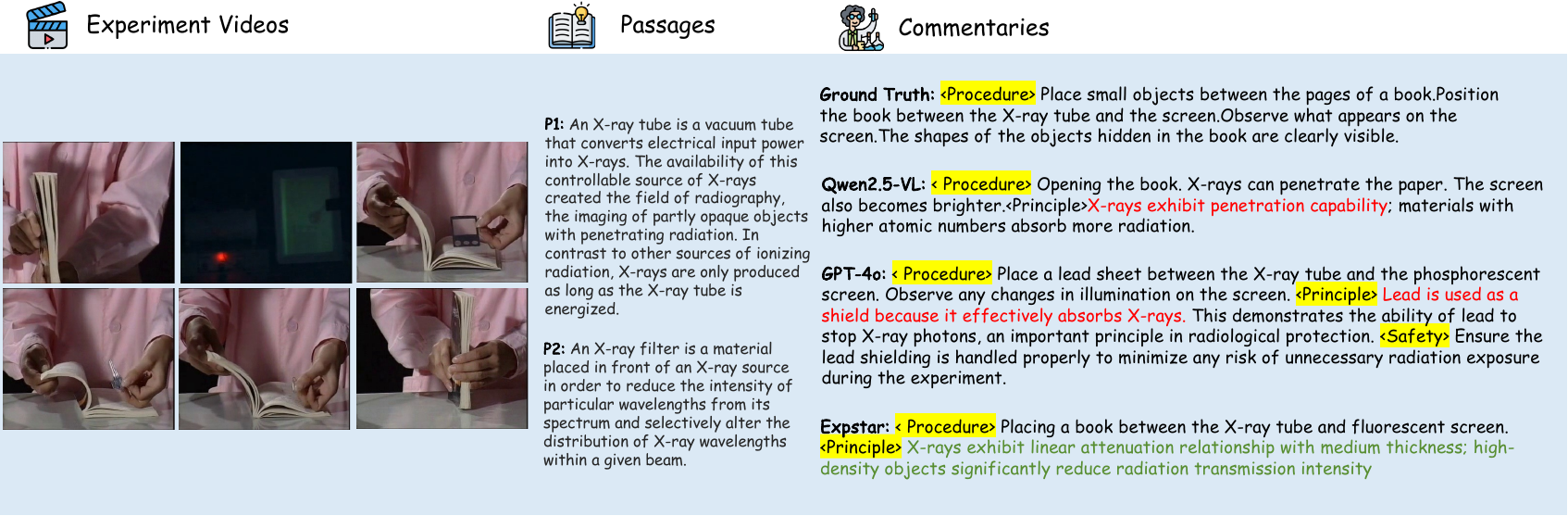}
    \caption{A case of the optics experiment. The \textcolor[RGB]{88,142,49}{green} and \textcolor[RGB]{255,0,0}{red} denote correct and incorrect prediction, respectively.}
    \label{fig:app_case1}
\end{figure*}

\begin{figure*}[h]
    \centering
    \includegraphics[width=1\linewidth]{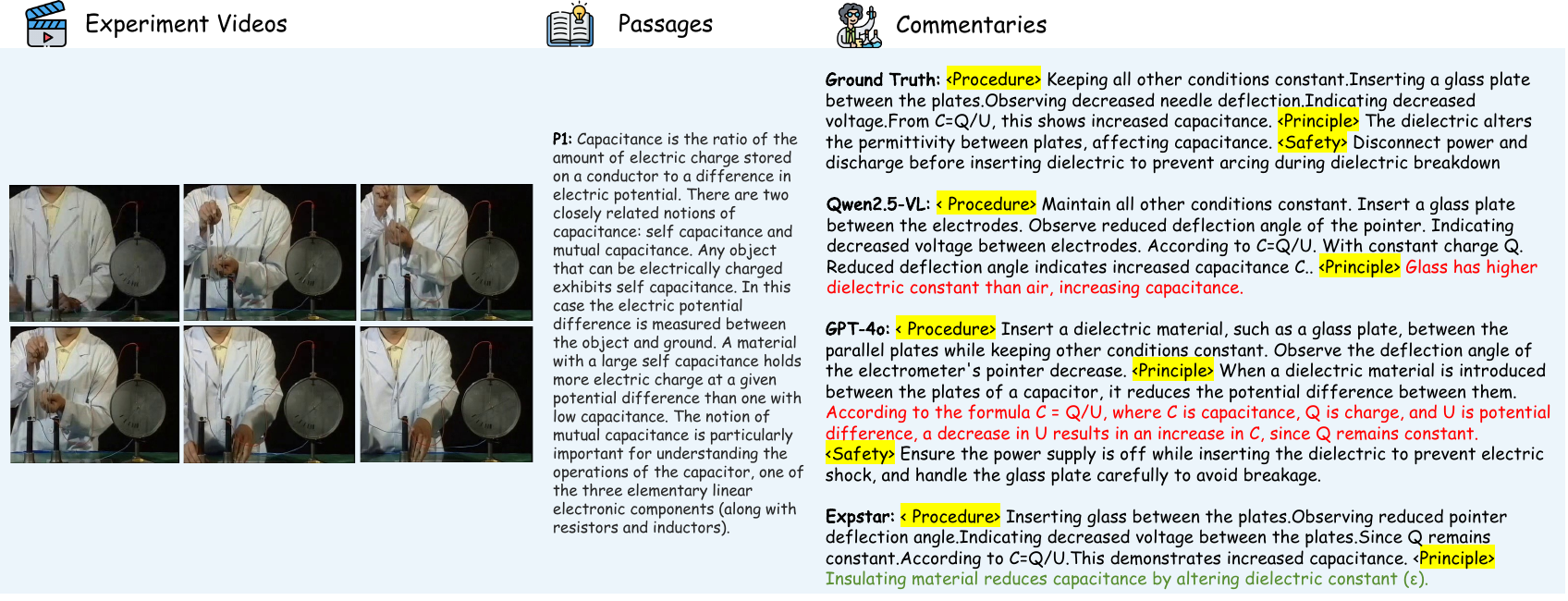}
    \caption{A case of the mechanics experiment.The \textcolor[RGB]{88,142,49}{green} and \textcolor[RGB]{255,0,0}{red} denote correct and incorrect prediction, respectively.}
    \label{fig:app_case2}
\end{figure*}

\begin{figure*}[h]
    \centering
    \includegraphics[width=1\linewidth]{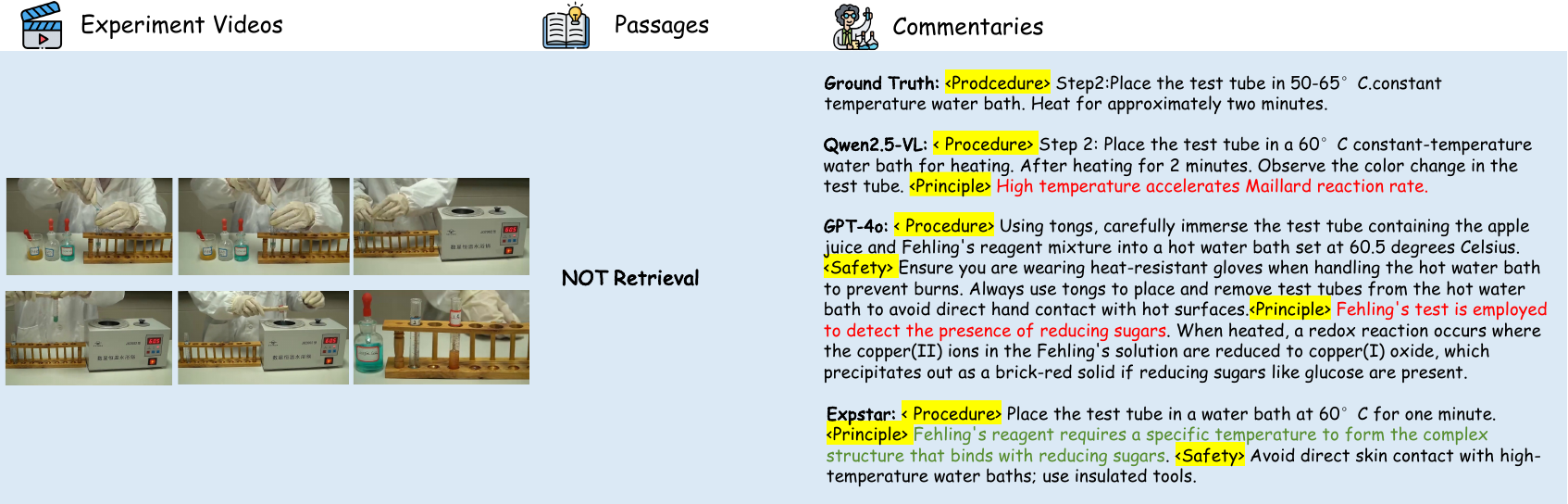}
    \caption{A case of biography experiment.The \textcolor[RGB]{88,142,49}{green} and \textcolor[RGB]{255,0,0}{red} denote correct and incorrect prediction, respectively.}
    \label{fig:app_case3}
\end{figure*}

\begin{figure*}[h]
    \centering
    \includegraphics[width=1\linewidth]{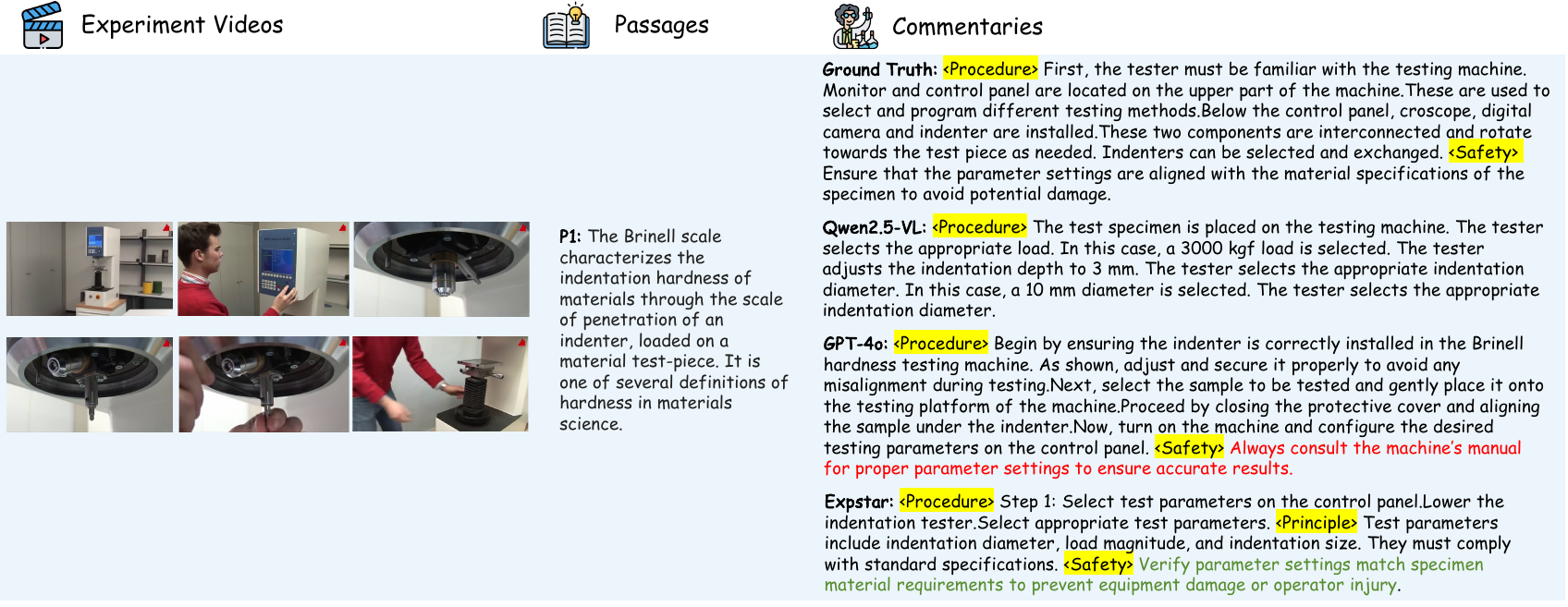}
    \caption{A case of the material engineering experiment. The \textcolor[RGB]{88,142,49}{green} and \textcolor[RGB]{255,0,0}{red} denote correct and incorrect prediction, respectively.}
    \label{fig:app_case4}
\end{figure*}

\begin{figure*}[h]
    \centering
    \includegraphics[width=1\linewidth]{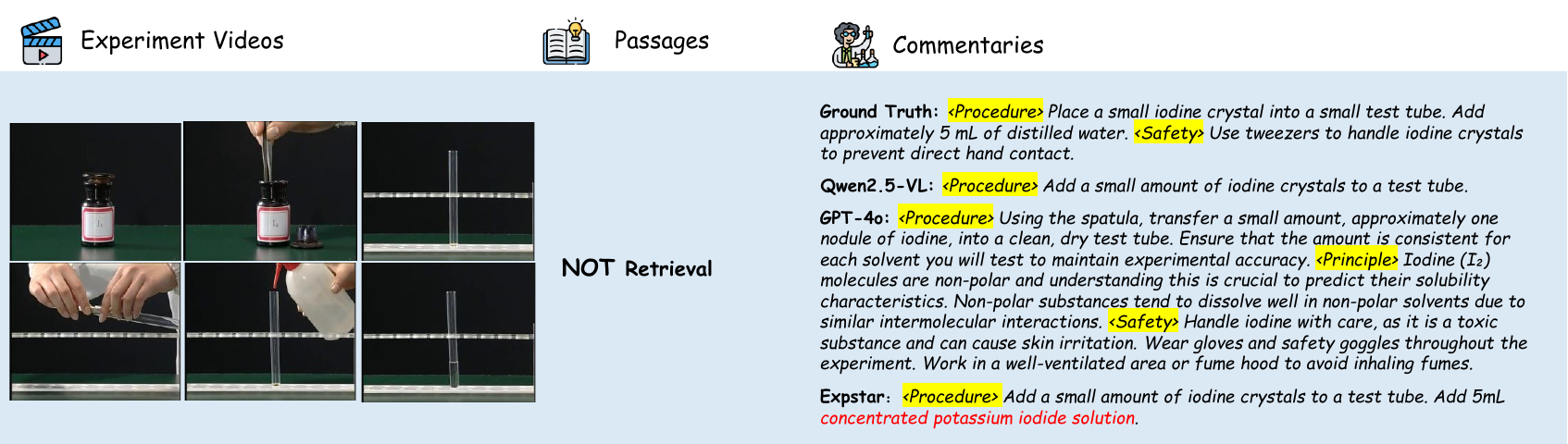}
    \caption{A failure case. The \textcolor[RGB]{88,142,49}{green} and \textcolor[RGB]{255,0,0}{red} denote correct and incorrect prediction, respectively.}
    \label{fig:bad_case}
\end{figure*}

Figures \ref{fig:app_case1}–\ref{fig:app_case4} present representative cases from our experiments, covering a range of disciplines, such as optics and chemistry. In contrast, Figure \ref{fig:bad_case} illustrates a failure case observed during the experiment.

\end{document}